\documentclass{article}
\usepackage{amsmath,amssymb,graphicx,relsize,mathtools}
\usepackage{hyperref}
\usepackage{booktabs} 
\usepackage{svg}
\usepackage{cleveref}

\newcommand{\R}{\mathbb{R}}
\newcommand{\I}{\mathbf{I}}

\title{On Geometry Regularization in Autoencoder Reduced-Order Models with Latent Neural ODE Dynamics}
\author{Mikhail Osipov\\
{Independent Researcher, Italy \footnote{osipov.ma@phystech.edu}}
}
\date{March 3, 2026}

\begin{document}

\maketitle

\begin{abstract}
We investigate geometric regularization strategies for learned latent representations in encoder--decoder reduced-order models. In a fixed experimental setting for the advection--diffusion--reaction (ADR) equation, we model latent dynamics using a neural ODE and evaluate four regularization approaches applied during autoencoder pre-training: (a) near-isometry regularization of the decoder Jacobian, (b) a stochastic decoder gain penalty based on random directional gains, (c) a second-order directional curvature penalty, and (d) Stiefel projection of the first decoder layer. Across multiple seeds, we find that (a)--(c) often produce latent representations that make subsequent latent-dynamics training with a frozen autoencoder more difficult, especially for long-horizon rollouts, even when they improve local decoder smoothness or related sensitivity proxies. In contrast, (d) consistently improves conditioning-related diagnostics of the learned latent dynamics and tends to yield better rollout performance. We discuss the hypothesis that, in this setting, the downstream impact of latent-geometry mismatch outweighs the benefits of improved decoder smoothness.
\end{abstract}

\section{Introduction}

Autoencoder (AE) architectures are widely used in machine learning (ML) and scientific machine learning (SciML). 
Dimensionality reduction via an encoder--decoder construction enables the modeling of high-dimensional dynamics in a low-dimensional latent space~\cite{LEE2020108973, Fulton2019, Fresca2021, FRESCA2022114181, Fries_2022}.

Neural ordinary differential equations (NODE)~\cite{NEURIPS2018_69386f6b, kidger2022neuraldifferentialequations} are commonly employed to learn continuous-time dynamics in latent space. 
More broadly, operator-learning methods have also been successfully applied to partial differential equations (PDEs)~\cite{JMLR:v24:21-1524}.

Recent works~\cite{ballini2025modelreductionparametricordinary,farenga2024latentdynamicslearningnonlinear} provide mathematical frameworks for reduced-order modeling (ROM) via autoencoders, discuss theoretical limitations, and demonstrate computational efficiency in representative benchmarks.

A fundamental concern in autoencoder-based ROM is that the latent space $\R^d$ typically has lower dimension than the ambient space $\R^n$. 
When $d<n$, the encoder $E:\R^n \to \R^d$ cannot be globally injective and therefore introduces information loss away from the data manifold. 
Consequently, the decoder $D:\R^d \to \R^n$ may exhibit locally expansive behavior along certain directions of the manifold. 
This can result in amplification of latent-space errors after decoding, an effect that becomes particularly pronounced in long-horizon rollouts.

The amplification of latent errors by the decoder can be quantified using its Lipschitz constant~\cite{Brivio2024}. 
Let $u(t)$ denote the true (discretized) trajectory in the ambient space and let $z(t)=E(u(t))$ be its latent representation. 
Let $\hat z(t)$ be a numerical approximation in latent space and define $\hat u(t)=D(\hat z(t))$. 
Then
\[
\|u - \hat u\|_2
= \|u - D(\hat z)\|_2
\le
\|u - D(E(u))\|_2
+
\|D(E(u)) - D(\hat z)\|_2
\le
\varepsilon_{\text{recon}}
+
L_D \|z - \hat z\|_2,
\]
where $\varepsilon_{\text{recon}}$ denotes the reconstruction error and $L_D$ is a local Lipschitz constant of the decoder on the relevant region of latent space.

One classical approach to controlling sensitivity in representation learning is to introduce Jacobian-based regularization~\cite{Rifai2011ContractiveAE, Ghosh2020From, Kumar2020RegularizedAV, lee2022regularized, pmlr-v202-nazari23a}. Some authors also discuss curvature regularization of autoencoders~\cite{Kim2024VAECR}. Another commonly used way of regularization is related to Stiefel projection~\cite{Higham1986PolarDecomposition, Absil2008MatrixManifolds, Cisse2017, Huang2017, DBLP:journals/corr/abs-2002-01113}.

In this work, we investigate several heuristics aimed at controlling the decoder's local sensitivity and geometric properties, as well as a structured projection onto the Stiefel manifold:
\begin{itemize}
    \item[(a)] a local isometry penalty on the decoder Jacobian, \\
    enforcing $\|J_D^\top J_D - \I\|_{\text{Frobenius}} \approx 0$;
    \item[(b)] a stochastic decoder gain penalty based on directional norms $\|J_D v\|$ for random unit vectors $v$;
    \item[(c)] a second-order directional curvature penalty based on variations of Jacobian--vector products;
    \item[(d)] a Stiefel (orthonormal-column) projection applied to the first decoder layer.
\end{itemize}

We demonstrate numerically, in the setting of an advection--diffusion--reaction (ADR) system combined with a latent neural ODE, that pretraining the autoencoder with regularizers (a)--(c) often produces latent geometries for which learning stable latent dynamics becomes more difficult and long-horizon rollouts degrade. 
This occurs even when intrinsic geometric diagnostics indicate improved local smoothness or reduced decoder-sensitivity proxies. 
In contrast, the Stiefel projection (d) consistently improves conditioning-related diagnostics of the learned latent dynamics and yields slightly better (or at least not worse) rollout performance.

\section{Geometry Regularizers}
\label{sec:geometry}
In this section, we describe the regularization methods used in our experiments and provide brief motivation for each of them. Full formal definitions are given in Appendix~\ref{app:geometry}.

\subsection{(a) Near-isometry Jacobian penalty}
\label{sec:iso}

A classical approach to autoencoder regularization is based on Jacobian control~\cite{Rifai2011ContractiveAE}. 
As one of our regularization options, we use a decoder-Jacobian near-isometry penalty, added to the loss during autoencoder pre-training:
\[
\mathcal{R}_a
\;=\;
\mathbb{E}\!\left[\left\|J_D^\top J_D - \I\right\|_F^2\right].
\]
In practice, this term can be evaluated either exactly (via basis Jacobian--vector products) or approximately using a Hutchinson-style stochastic estimator~\cite{Hutchinson1990TraceEstimator}. 
In our experiments, we did not observe substantial differences between the exact and stochastic evaluations of this penalty.

\subsection{(b) Directional gain penalty}
\label{sec:gain}

To control decoder sensitivity without enforcing full near-isometry, we also consider a weaker directional gain penalty:
\[
\mathcal{R}_b
\;=\;
\mathbb{E}_v\!\left[\bigl(\|J_D v\| - 1\bigr)^2\right],
\]
where $v$ is a random unit vector in latent space. 
This penalty encourages approximately unit gain along sampled directions, but unlike \(\mathcal{R}_a\), it does not directly constrain the full spectrum of \(J_D\). 
The required Jacobian--vector products are estimated numerically using finite differences.

\subsection{(c) Flatness penalty}
\label{sec:curvature}

To probe the local flatness of the decoder-induced geometry without explicitly enforcing strong isometry or gain constraints, we introduce a curvature-related penalty:
\[
\mathcal{R}_c
\;\approx\;
\mathbb{E}\!\left[\left\|\nabla^2 D(z)\right\|^2\right]
\;\approx\;
\mathbb{E}\!\left[
\frac{\left\|J_D(z+\varepsilon v)\,v - J_D(z)\,v\right\|^2}{\varepsilon^2}
\right].
\]
This term approximates a directional second-order variation of the decoder. To compute it, we use autodiff JVPs and estimate directional curvature by finite-differencing these JVPs across a shifted basepoint.

\subsection{(d) Stiefel projection}
\label{sec:stiefel}

Projection onto the Stiefel manifold is a standard tool in optimization on matrix manifolds~\cite{Higham1986PolarDecomposition, Absil2008MatrixManifolds, Cisse2017, Huang2017, DBLP:journals/corr/abs-2002-01113}. 
In our setting, after each gradient step, the weight matrix of a selected decoder layer is projected back onto the Stiefel manifold, enforcing orthonormal columns for that layer.

Unlike the penalties above, this method imposes a structural constraint only on a single layer (in our case, the first convolutional layer of the decoder). 
As a result, it can improve the conditioning of that layer without directly enforcing near-isometry, low curvature, or controlled gain for the full decoder Jacobian. 
Therefore, it should be viewed as a partial architectural regularization rather than as an explicit constraint on the geometry of the decoder as a whole.

\section{Numerical experiments}

\subsection{Dataset generation}

We generate full-order trajectories for a parametric advection--diffusion--reaction (ADR) problem using a \texttt{dolfinx} finite-element solver on the unit square. 
The spatial discretization uses first-order Lagrange elements on a uniform triangular mesh with $31\times 31$ subdivisions, yielding $32\times 32=1024$ scalar degrees of freedom per snapshot.

The governing dynamics depend on a parameter vector $\mu=(\mu_1,\mu_2,\mu_3)\in\R^3$, where $\mu_1$ controls diffusion and $(\mu_2,\mu_3)$ determine the center of a Gaussian source term. 
The reaction coefficient is fixed, and the advection field is time-dependent.

Full-order trajectories are integrated on the interval $[0,T]$, with $T=10\pi$, using backward Euler with $1000$ time steps. 
This produces $1001$ fine-grid snapshots per trajectory. 
For reduced-model training, we additionally construct a coarse temporal grid by retaining every fifth full-order step, resulting in $201$ snapshots per trajectory.

The parameter domain is split into three subsets: a structured training / validation set inside an inner parameter box, an interpolation test set formed by midpoint parameters within the same box, and an extrapolation test set sampled outside the training box but within a larger outer box. 

We further use the times $T_1=4\pi$ and $T_2=5\pi$ to distinguish the training, interpolation-in-time, and extrapolation-in-time portions of each trajectory.

Our protocol generally follows the approach of~\cite{farenga2024latentdynamicslearningnonlinear}; more detailed description of the dataset generation protocol is provided in Appendix~\ref{app:dataset_generation}.

\subsection{Training}

We first pre-train a family of autoencoders on individual snapshots from the training set, applying one of the regularization methods described in Sections~\ref{sec:iso}--\ref{sec:stiefel}. 
As a baseline, we also train an unregularized encoder--decoder pair.

All autoencoders use the same architecture and differ only in the geometry regularization strategy applied during pre-training. 
To reduce the possibility that the downstream results are explained by a particularly favorable realization of one autoencoder, we train multiple independent autoencoder runs for each method (in our main setup, two AE seeds per method). 
Each pre-training run is performed for the same computational budget of $200$ epochs.

For checkpoint selection, our main protocol uses a shared validation-MSE target that is reached by all methods; for each autoencoder, we select the first epoch at which this target is attained. 
In a separate series of experiments, we instead use method-specific checkpoint selection, choosing for each method the epoch with the lowest validation loss.

After autoencoder pre-training, we freeze the encoder and decoder and train a neural ODE in latent space. 
For each selected autoencoder, we train multiple independent neural ODE runs (in our main setup, $16$ NODE seeds per method). 
Crucially, for a given NODE seed, all methods are initialized from the same latent-dynamics initialization, while each method uses its own frozen autoencoder. 
This makes the neural-ODE comparison paired at the seed level and allows us to attribute differences more directly to the latent geometry induced by the corresponding autoencoder, rather than to a favorable initialization of the latent dynamics model.

This two-stage protocol isolates the effect of the autoencoder-induced latent geometry on the learnability of the latent dynamics and on long-horizon rollout quality.

\subsection{Evaluation}

For evaluation, we sample a fixed set of rollout windows from the chosen dataset split and use exactly the same windows for all methods, so that comparisons are paired at the window level. 
In our main setup, evaluation is performed on the fine temporal grid for several rollout horizons.

For each sampled window, we encode the initial state, integrate the latent neural ODE on the corresponding absolute-time grid, decode the predicted latent trajectory, and compare it with the full-order trajectory. 
We report absolute and relative rollout errors aggregated over time, including mean and maximum errors.

In addition, on a subset of evaluation windows we compute intrinsic diagnostics of the learned latent dynamics and decoder, including the singular values and condition number of the latent-dynamics Jacobian, a decoder-gain proxy, and the latent tracking error.

Because the dominant variability in our experiments arises from neural-ODE training rather than from the choice of evaluation windows, our primary comparisons are performed at the level of matched NODE seeds. 
That is, for a fixed NODE seed, all methods are compared using the same rollout windows and the same latent-dynamics initialization, so the resulting differences can be interpreted as seed-wise paired differences between methods. 
Window-level pairing is therefore used to reduce evaluation noise within each run, while the main statistical comparison is based on aggregating and comparing outcomes across matched NODE seeds.

All summary statistics are reported per method, and paired comparisons against the unregularized baseline are performed using these matched per-seed evaluations.

We provide detailed formal evaluation protocol in Appendix~\ref{app:eval_protocol}.

\section{Results}
\subsection{Training dynamics}

As expected, adding geometry-based objectives during autoencoder pre-training affects optimization and convergence. 
To ensure a fair comparison between the regularized methods and the unregularized baseline, we use a shared checkpoint-selection criterion during autoencoder pre-training: for each autoencoder seed (versions $=1,2$), we define a common validation reconstruction target (validation MSE) that is reached by all methods, and we select, for each method, the first checkpoint at which this target is attained. 
Therefore, up to small differences, all methods enter the NODE stage with comparable autoencoder reconstruction quality.

\begin{table}[h]
\caption{Latent NODE training metadata for different regularization methods}
\label{tab:metadata}
\smallskip
\begin{tabular}{lrrrr}
\toprule
Method & $\mathcal{L}_{\text{val}}$ & $\mathrm{CI}_{95}(\mathcal{L}_{\text{val}})$ & $ep_{\text{target}}$ & $\mathrm{CI}_{95}(ep_{\text{target}})$ \\
\midrule
Stiefel projection & $8.6\cdot 10^{-4}$ & $1.0\cdot 10^{-4}$ & 13.5 & 1.6 \\
Vanilla baseline   & $8.8\cdot 10^{-4}$ & $1.2\cdot 10^{-4}$ & 14.1 & 1.4 \\
Curvature penalty  & $1.1\cdot 10^{-3}$ & $1.4\cdot 10^{-4}$ & 14.5 & 1.2 \\
Gain penalty       & $1.4\cdot 10^{-3}$ & $1.8\cdot 10^{-4}$ & 18.1 & 2.5 \\
Isometry penalty   & $1.4\cdot 10^{-3}$ & $1.8\cdot 10^{-4}$ & 19.3 & 3.4 \\
\bottomrule
\end{tabular}

\smallskip
\textit{For each method, the table reports the mean best validation loss $\mathcal{L}_{\text{val}}$, the corresponding 95\% confidence-interval half-width, the mean epoch $ep_{\text{target}}$ at which the shared validation target is first reached, and the corresponding 95\% confidence-interval half-width. Values are averaged across matched AE/NODE realizations.}
\end{table}

Despite this matching at the autoencoder level, the subsequent NODE training dynamics differ substantially across methods. 
For the latent neural ODE, we monitor: (i) the best validation loss (trajectory-rollout MSE on the validation split, with fixed horizon $H=40$ on the coarse grid), (ii) the epoch at which this best validation loss is achieved, and (iii) the epoch at which a shared validation-loss target is first reached in a post-hoc comparison (see Table~\ref{tab:metadata}).

With a fixed training budget of 50 epochs and a frozen autoencoder, NODEs built on top of the Stiefel-regularized autoencoder converge to a slightly better validation loss than the vanilla baseline. 
The curvature-regularized AE leads to a moderately worse validation loss, while the isometry- and gain-regularized autoencoders lead to substantially worse validation losses. 
A similar pattern is visible in the shared-target analysis: Stiefel-regularized, baseline, and curvature-regularized systems reach the shared validation target at comparable epochs (with the same qualitative ordering), whereas the isometry and gain variants reach the same target substantially later.

We interpret these differences as an early indication that the isometry and gain penalties produce latent representations that are less favorable for learning stable latent dynamics, even when the corresponding autoencoders are matched to the baseline in reconstruction error.

\subsection{Rollout dynamics}

We evaluate rollout quality on the fine temporal grid and on the extrapolation split for several horizons ($H=80,160,240,320$). 
In the main experimental protocol, autoencoder checkpoints are selected using the shared reconstruction target described above, while NODE checkpoints are selected at the epoch with the best validation loss. 
For the NODE weights, we additionally apply stochastic weight averaging (SWA) over the three checkpoints corresponding to the selected epoch and the two preceding epochs.

\begin{figure}[ht!]
\centering
\includegraphics[width=80mm]{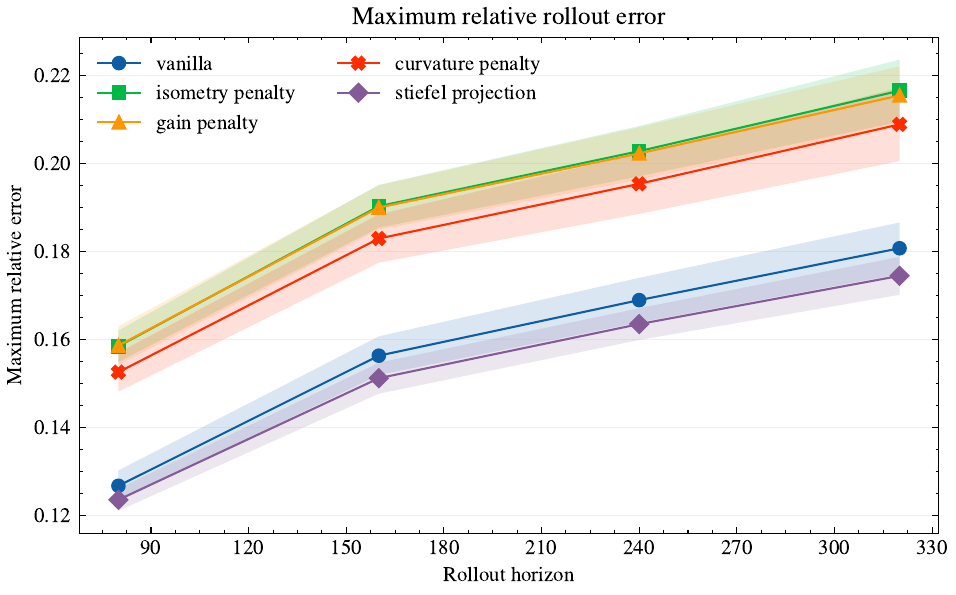}
\caption{\textit{Maximum relative error over rollout horizon values $\in\{80,160,240,320\}$ on fine grid for different regularization methods and "vanilla" baseline with no regularization. Uncertainty is reported as a 95\% confidence interval.}}
\label{fig:rollout_max}
\end{figure}

\begin{figure}[ht!]
\centering
\includegraphics[width=80mm]{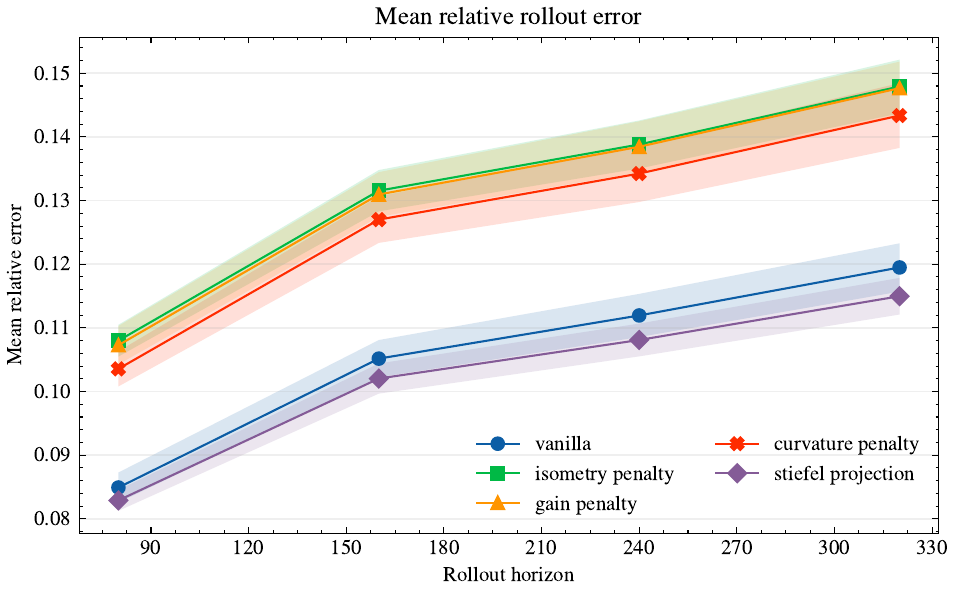}
\caption{\textit{Mean relative error over rollout horizon values $\in\{80,160,240,320\}$ on fine grid for different regularization methods and "vanilla" baseline with no regularization. Uncertainty is reported as a 95\% confidence interval.}}
\label{fig:rollout_mean}
\end{figure}

Across all tested horizons, the same qualitative pattern is observed. 
NODEs trained on frozen autoencoders produced by the isometry, gain, and curvature regularizers exhibit larger mean relative rollout errors and larger maximum relative rollout errors than the vanilla baseline. 
By contrast, the Stiefel variant is the only method that is consistently competitive with, and typically slightly better than, the vanilla model. 
This trend is visible in the horizon-dependent plots of $\varepsilon_{\text{mean}}$ and $\varepsilon_{\text{max}}$ (Figures~\ref{fig:rollout_max}--\ref{fig:rollout_mean}).

To make the comparison more robust, we also compute paired differences between each method and the vanilla baseline. 
For each autoencoder seed and NODE seed, we first average rollout errors over the common set of evaluation windows and then compute the paired difference
\[
\Delta = \varepsilon(\texttt{vanilla}) - \varepsilon(\texttt{method}).
\]
Thus, positive values indicate that the regularized method improves over the baseline. 
Averaged over all matched autoencoder/NODE realizations, Stiefel is the only method with a positive paired effect, while all other methods show negative paired differences for both mean and maximum relative rollout error. 
In particular, the Stiefel improvement remains statistically consistent in paired one-sided tests, whereas curvature, gain, and isometry regularization methods are uniformly worse than vanilla (see Table~\ref{tab:paired}).

\begin{table}[h]
\caption{Paired rollout-error differences relative to the vanilla baseline}
\label{tab:paired}
\vspace{5pt}
\begin{tabular}{lrr}
\toprule
Method & (A) Paired $\Delta \varepsilon_{\text{max}}$ & (B) Paired $\Delta \varepsilon_{\text{mean}}$ \\
\midrule
Stiefel projection & 0.006 $\pm 4.2 \cdot 10^{-3}$ & 0.005 $\pm 3.1 \cdot 10^{-3}$ \\
Curvature penalty  & -0.028 $\pm 8.7 \cdot 10^{-3}$ & -0.024 $\pm 5.1 \cdot 10^{-3}$ \\
Gain penalty       & -0.035 $\pm 7.6 \cdot 10^{-3}$ & -0.028 $\pm 4.6 \cdot 10^{-3}$ \\
Isometry penalty   & -0.036 $\pm 7.9 \cdot 10^{-3}$ & -0.028 $\pm 4.5 \cdot 10^{-3}$ \\
\bottomrule
\end{tabular}

\smallskip
\textit{Paired differences are defined as $\Delta = \varepsilon(\texttt{vanilla})-\varepsilon(\texttt{method})$, so positive values indicate an improvement over the unregularized baseline. 
Column (A) reports paired differences in maximum relative rollout error, and column (B) reports paired differences in mean relative rollout error. 
Values are averaged over matched AE/NODE realizations ($2\times16$ pairs). 
Results are shown for rollout horizon $H=320$ on the fine grid, with NODE checkpoints selected by the best validation error. 
Values are reported as mean $\pm 95\%$ confidence interval half-width.}
\end{table}

Our interpretation is that, even when the autoencoders are matched in reconstruction quality, regularizers (a)--(c) favor latent representations that make latent-NODE training harder and degrade long-horizon rollouts.

In the appendix, we additionally report a stricter comparison in which not only the AE checkpoint, but also the NODE checkpoint, is selected using a shared validation-rollout target (defined within each autoencoder seed). 
The same ordering persists: even when methods are matched by validation rollout quality, they still diverge on extrapolation, and this additional matching does not rescue the curvature, gain, or isometry regularization variants.

\subsection{Latent dynamics}

To support the hypothesis that some regularizers induce a \emph{hostile latent geometry}, we report intrinsic diagnostics of the learned latent dynamics and decoder.

In Table~\ref{tab:conditioning}, we summarize three quantities: (i) the conditioning of the learned latent-dynamics Jacobian (not the decoder Jacobian), (ii) a decoder-gain proxy, and (iii) the latent tracking error. 
For the conditioning statistic, we use the run-level median (50th percentile) of the condition number and then average this quantity across runs; this avoids undue sensitivity to extreme outliers while preserving the ordering between methods.

\begin{table}[h]
\caption{Conditioning of the learned latent dynamics, decoder gain, latent error}
\label{tab:conditioning}
\vspace{5pt}
\begin{tabular}{lrrr}
\toprule
Method & Dyn. conditioning & Decoder gain & Latent error \\
\midrule
Stiefel projection & 124.2 $\pm 11.5$ & 4.428 $\pm 9.7\cdot 10^{-2}$ & 0.988 $\pm 0.022$ \\
Vanilla            & 145.3 $\pm 13.9$ & 3.380 $\pm 3.0\cdot 10^{-3}$ & 1.156 $\pm 0.043$ \\
Curvature penalty  & 316.6 $\pm 18.9$ & 1.570 $\pm 1.6\cdot 10^{-2}$ & 3.434 $\pm 0.119$ \\
Isometry penalty   & 407.2 $\pm 19.8$ & 1.000 $\pm 7.0\cdot 10^{-5}$ & 4.066 $\pm 0.122$ \\
Gain penalty       & 424.3 $\pm 21.6$ & 1.017 $\pm 7.2\cdot 10^{-3}$ & 4.217 $\pm 0.133$ \\
\bottomrule
\end{tabular}

\smallskip
\textit{Summary of intrinsic diagnostics at rollout horizon $H=320$ on the fine grid. 
The first column reports the median (50th percentile) of the condition number of the learned latent-dynamics Jacobian, averaged over matched AE/NODE realizations. 
The second and third columns report the decoder-gain proxy and latent tracking error, respectively. 
NODE checkpoints are selected by the best validation error. 
Values are reported as mean $\pm 95\%$ confidence interval half-width.}
\end{table}

The Stiefel-regularized autoencoder yields better-conditioned latent dynamics and lower latent error than the vanilla baseline, despite exhibiting a slightly larger decoder gain. 
By contrast, the curvature, isometry, and gain regularizers substantially reduce decoder gain---indeed, almost perfectly so for the isometry and gain penalties---but this does not translate into better rollouts. 
Instead, these methods produce worse conditioning of the latent dynamics Jacobian and substantially larger latent errors. 
The curvature penalty effect is intermediate in this respect, but still clearly underperforms the vanilla baseline in rollout quality.

Taken together, these diagnostics support the conclusion that reducing decoder-side expansion alone is not sufficient in this setting: the structure of the latent representation, and its effect on the learnability and conditioning of the latent dynamics, is more consequential.

\section{Conclusion}

We presented a systematic study of geometry-based regularization methods in a controlled latent-ROM setting combining an ADR benchmark, a convolutional autoencoder, and a latent neural ODE with decoupled autoencoder/NODE training.

Although Jacobian-based regularization is a natural approach for controlling decoder sensitivity, our experiments show that the decoder-side regularizers considered here---near-isometry, directional gain, and curvature---consistently worsen downstream latent-NODE performance relative to the unregularized baseline. 
This degradation appears both in NODE training dynamics and, more importantly, in long-horizon rollout quality.

Our results support the hypothesis that, in this setting, regularizers (a)--(c) induce latent representations that are less suitable for learning stable continuous-time latent dynamics. 
Any reduction in decoder-side error amplification achieved by these regularizers does not compensate for the accompanying deterioration in latent-dynamics conditioning and latent tracking accuracy.

In contrast, the Stiefel projection---even when applied only to a single decoder layer---consistently improves conditioning-related latent-dynamics diagnostics and yields slightly better rollout performance on average. 
This suggests that milder structural constraints may be more effective than direct Jacobian-based penalties when the goal is not only reconstruction quality, but also stable and accurate latent-dynamics learning.

We hope these findings are useful for future work on latent ROMs for ADR-type systems, on geometry-aware regularization more broadly, and on more practical strategies such as mixed regularizers, conditioning-aware penalties, and joint training with an unfrozen autoencoder.

We provide full implementation code available at \url{https://github.com/miosipof/adr_latent_dynamics}.



\appendix

\section{Formal definition of losses and geometry regularizers}
\label{app:geometry}

\subsection{Summary of the regularizers}

\begin{itemize}
    \item \textbf{Near-isometry penalty:} enforces $J_D(z)^\top J_D(z)\approx \I$, i.e., local near-isometry of the decoder via stochastic Hutchinson evaluation. Used with weight $= 0.1$.
    \item \textbf{Directional gain penalty:} matches typical values of $\|J_D(z)v\|$ to a target over random directions; this controls average Jacobian scale rather than worst-case expansion. Used with weight $= 0.1$.
    \item \textbf{Directional curvature penalty:} penalizes changes in $J_D(z)v$ along the same direction $v$, i.e., directional second-order variation of the decoder.  Used with weight $= 1.0$.
    \item \textbf{Stiefel projection:} enforces column-orthonormality of the first decoder-layer weight matrix, acting as a partial structural regularizer rather than a direct constraint on the full decoder Jacobian.
\end{itemize}

\subsection{Notation and tensor shapes}

Let
\begin{itemize}
    \item $u \in \R^{1\times 32\times 32}$ denote a normalized scalar-field snapshot;
    \item $z \in \R^{1\times 4\times 4}$ denote a latent field;
    \item $E_\phi:\R^{1\times 32\times 32}\to \R^{1\times 4\times 4}$ be the encoder;
    \item $D_\psi:\R^{1\times 4\times 4}\to \R^{1\times 32\times 32}$ be the decoder.
\end{itemize}
We denote by
\[
d \coloneqq 1\cdot 4\cdot 4 = 16,
\qquad
n \coloneqq 1\cdot 32\cdot 32 = 1024
\]
the latent and ambient dimensions after flattening, respectively. We write $\mathrm{vec}(\cdot)$ for vectorization, so that
\[
\mathrm{vec}(z)\in \R^d,
\qquad
\mathrm{vec}(D_\psi(z))\in \R^n.
\]

\paragraph{Normalization anchor.}
We also use the constant field
\[
u_{\min} \coloneqq -\mathbf{1}\in \R^{1\times 32\times 32},
\]
which corresponds to the physical minimum field value after min--max normalization to $[-1,1]$.

\subsection{Autoencoder architecture}

Both encoder and decoder are spatially coherent convolutional networks with residual blocks and bilinear down/up-sampling:
\[
E_\phi
=
\mathrm{Conv}_{\mathrm{out}}
\circ
\mathrm{Down}^{(L)}
\circ
\mathrm{Conv}_{\mathrm{in}},
\qquad
D_\psi
=
\mathrm{Conv}_{\mathrm{out}}
\circ
\mathrm{Up}^{(L)}
\circ
\mathrm{Conv}_{\mathrm{in}}.
\]
Each down/up stage consists of a pre-activation residual block followed by bilinear interpolation by a factor $1/2$ (down) or $2$ (up), repeated $L$ times. In our experiments, $L=3$, corresponding to the resolution sequence
\[
32 \;\to\; 16 \;\to\; 8 \;\to\; 4.
\]

\subsection{Autoencoder training objective}

Given a minibatch $\{u^{(b)}\}_{b=1}^B$, define
\[
z^{(b)} = E_\phi(u^{(b)}),
\qquad
\hat u^{(b)} = D_\psi(z^{(b)}).
\]

\paragraph{Reconstruction loss.}
The batch reconstruction loss is
\[
\mathcal{L}_{\mathrm{mse}}(\phi,\psi)
=
\frac{1}{B}\sum_{b=1}^B
\frac{1}{n}
\left\|
\mathrm{vec}\bigl(\hat u^{(b)}\bigr)
-
\mathrm{vec}\bigl(u^{(b)}\bigr)
\right\|_2^2.
\]

\paragraph{Anchor loss.}
We additionally penalize reconstruction of the constant minimum field:
\[
\mathcal{L}_{\mathrm{anc}}(\phi,\psi)
=
\frac{1}{n}
\left\|
\mathrm{vec}\bigl(D_\psi(E_\phi(u_{\min}))\bigr)
-
\mathrm{vec}(u_{\min})
\right\|_2^2.
\]

\paragraph{Regularized AE objective.}
Let $\mathcal{R}(D_\psi; z)$ denote a decoder-side regularizer evaluated at a latent batch $z$ (defined in Section~\ref{app:regs}). Autoencoder pre-training uses
\[
\mathcal{L}_{\mathrm{AE}}(\phi,\psi)
=
\mathcal{L}_{\mathrm{mse}}(\phi,\psi)
+
\lambda_{\mathrm{reg}}(e)\,\mathcal{R}(D_\psi; z)
+
w_{\mathrm{anc}}\,\mathcal{L}_{\mathrm{anc}}(\phi,\psi),
\]
where $e$ is the epoch index and $\lambda_{\mathrm{reg}}(e)$ is linearly ramped after a warmup period:
\[
\lambda_{\mathrm{reg}}(e)
=
\begin{cases}
0, & e \le e_{\mathrm{warm}}, \\[4pt]
\lambda_{\mathrm{reg}}
\displaystyle
\frac{e-e_{\mathrm{warm}}}{E-e_{\mathrm{warm}}}, & e_{\mathrm{warm}} < e \le E.
\end{cases}
\]

\paragraph{Stiefel projection during AE training.}
For the Stiefel variant, and only after warmup, the weights of the first decoder convolution are projected after each optimizer step onto a Stiefel-type constraint (see Section~\ref{app:stiefel}).

\subsection{Latent NODE model and NODE training objective}

Let $u(t)\in \R^{1\times 32\times 32}$ denote a trajectory and let $\mu\in\R^{n_\mu}$ be the parameter vector (in our experiments, $n_\mu=3$). For a rollout window $(u_0,\dots,u_\ell)$ sampled at times $(t_0,\dots,t_\ell)$, we define
\[
z_0 = E_\phi(u_0),
\qquad
\dot z(t) = f_\theta(t,z(t),\mu),
\]
where the latent ODE is integrated by an explicit Runge--Kutta method (RK2 Ralston in our experiments). The decoded predictions are
\[
\hat u_i = D_\psi(z(t_i)),
\qquad
i=0,\dots,\ell.
\]

\paragraph{Rollout loss.}
The rollout loss is
\[
\mathcal{L}_{\mathrm{roll}}(\phi,\psi,\theta)
=
\frac{1}{B}\sum_{b=1}^B
\frac{1}{\ell+1}\sum_{i=0}^{\ell}
\frac{1}{n}
\left\|
\mathrm{vec}\bigl(\hat u_i^{(b)}\bigr)
-
\mathrm{vec}\bigl(u_i^{(b)}\bigr)
\right\|_2^2.
\]

\paragraph{Reconstruction-along-window loss.}
We also monitor reconstruction along the rollout window:
\[
\mathcal{L}_{\mathrm{rec}}(\phi,\psi)
=
\frac{1}{B}\sum_{b=1}^B
\frac{1}{\ell+1}\sum_{i=0}^{\ell}
\frac{1}{n}
\left\|
\mathrm{vec}\bigl(D_\psi(E_\phi(u_i^{(b)}))\bigr)
-
\mathrm{vec}\bigl(u_i^{(b)}\bigr)
\right\|_2^2.
\]

\paragraph{Anchor loss.}
The same anchor loss $\mathcal{L}_{\mathrm{anc}}$ is included during NODE training.

\paragraph{NODE training objective.}
Let $z_\star$ denote the latent state at which the decoder-side regularizer is evaluated (e.g., the last latent state in a window). The loss used for latent-dynamics training is
\[
\mathcal{L}_{\mathrm{NODE}}(\phi,\psi,\theta)
=
\mathcal{L}_{\mathrm{roll}}(\phi,\psi,\theta)
+
\lambda_{\mathrm{recon}}\,\mathcal{L}_{\mathrm{rec}}(\phi,\psi)
+
\lambda_{\mathrm{reg}}\,\mathcal{R}(D_\psi; z_\star)
+
w_{\mathrm{anc}}\,\mathcal{L}_{\mathrm{anc}}(\phi,\psi).
\]
When the autoencoder is frozen, only $\theta$ is optimized. When the autoencoder is unfrozen, $\phi$, $\psi$, and $\theta$ are optimized jointly (with a smaller learning rate for autoencoder parameters). In our main experiments we keep autoencder frozen during NODE training, so that effectively
\[
\mathcal{L}_{\mathrm{NODE}}(\phi,\psi,\theta) = \mathcal{L}_{\mathrm{roll}}(\phi,\psi,\theta).
\]

\subsection{Decoder-side regularizers}
\label{app:regs}

All regularizers are applied to the decoder $D_\psi$ and evaluated at a latent batch
\[
z \in \R^{B\times 1\times 4\times 4}.
\]
During AE pre-training, we use $z=E_\phi(u)$ for the current batch. During NODE training, we evaluate the regularizer at a selected latent state $z_\star$ from the current rollout window.

\subsubsection{Near-isometry penalty}

Define the Jacobian of the flattened decoder output with respect to the flattened latent state:
\[
J_{D_{\psi}}(z)
\coloneqq
\frac{\partial\, \mathrm{vec}(D_\psi(z))}{\partial\, \mathrm{vec}(z)}
\in \R^{n\times d}.
\]
Its pullback Gram matrix is
\[
G_\psi(z)
\coloneqq
J_{D_\psi}(z)^\top J_{D_\psi}(z)
\in \R^{d\times d}.
\]
The exact near-isometry penalty is
\[
\mathcal{R}_{\mathrm{a}}(D_\psi; z)
=
\frac{1}{B}\sum_{b=1}^B
\left\|
G_\psi\!\bigl(z^{(b)}\bigr)- \I_d
\right\|_F^2.
\]
This directly penalizes deviation of the decoder metric from the Euclidean metric in latent space.

\paragraph{Exact evaluation.}
In the exact variant, the full Jacobian is assembled by computing $J_{D_\psi}(z)e_i$ for the canonical basis vectors $e_i\in\R^d$ via forward-mode Jacobian--vector products (JVPs), and then forming $G_\psi(z)=J_{D_\psi}(z)^\top J_{D_\psi}(z)$.

\paragraph{Stochastic Hutchinson evaluation.}
Let
\[
A_\psi(z) \coloneqq J_{D_\psi}(z)^\top J_{D_\psi}(z) - \I_d.
\]
For any isotropic probe vector $v\in\R^d$ satisfying $\mathbb{E}[vv^\top]=\I_d$, one has
\[
\mathbb{E}_v\|A_\psi(z)v\|_2^2 = \|A_\psi(z)\|_F^2.
\]
Therefore the same quantity may be estimated stochastically as
\[
\mathcal{R}_a(D_\psi; z)
=
\frac{1}{B}\sum_{b=1}^B
\mathbb{E}_v
\left[
\left\|
\bigl(J_{D_\psi}(z^{(b)})^\top J_{D_\psi}(z^{(b)})-\I_d\bigr)v
\right\|_2^2
\right].
\]
In practice, this is approximated with a finite number of probe vectors using a JVP--VJP computation of $(J^\top J)v$, without explicitly forming $J$.

\paragraph{Detach convention.}
If the latent input $z$ is detached, the regularizer updates decoder parameters $\psi$ only. Otherwise, gradients may also flow upstream to the encoder and/or the latent dynamics model through the dependence of $z$ on those components.

\subsubsection{Directional gain penalty}

Our second penalty controls typical directional gain of the decoder Jacobian without enforcing full near-isometry:
\[
\mathcal{R}_{\mathrm{b}}(D_\psi; z)
=
\frac{1}{B}\sum_{b=1}^B
\mathbb{E}_v
\left[
\bigl(
\|J_{D_\psi}(z^{(b)})v\|_2 - \alpha
\bigr)^2
\right],
\]
where $v$ is a random unit vector in latent space and $\alpha$ is a target gain (equal to $1$ in our main experiments). This regularizer matches a typical directional gain and is related to the Frobenius scale of the Jacobian, but it does not control the full singular-value spectrum of $J_{D_\psi}(z)$.

\paragraph{Finite-difference JVP approximation.}
The directional derivative is approximated by a one-sided finite difference,
\[
J_{D_\psi}(z)v
\;\approx\;
\frac{
\mathrm{vec}(D_\psi(z+\varepsilon v))
-
\mathrm{vec}(D_\psi(z))
}{\varepsilon},
\]
with $\varepsilon>0$ a small finite-difference step.

\subsubsection{Directional curvature penalty}

To probe nonlinearity of the decoder along sampled directions, we use a curvature-related penalty based on changes of the directional derivative:

\[
\mathcal{R}_{\mathrm{c}}(D_\psi; z)
=
\frac{1}{B}\sum_{b=1}^B
\mathbb{E}_v
\left[
\left\|
\frac{
J_{D_\psi}(z^{(b)}+\varepsilon v)v
-
J_{D_\psi}(z^{(b)})v
}{\varepsilon}
\right\|_2^2
\right].
\]
The increment under the norm approximates a second directional derivative,
\[
\frac{J_{D_\psi}(z+\varepsilon v)v - J_{D_\psi}(z)v}{\varepsilon}
\approx
\bigl(\nabla^2 \mathrm{vec}(D_\psi)(z)\bigr)[v,v],
\]
so this regularizer penalizes large directional curvature.

\paragraph{Implementation detail.}
In contrast to the directional gain penalty, the JVPs in this term are computed exactly (up to autodiff accuracy) at both $z$ and $z+\varepsilon v$.

\subsection{Stiefel projection of the first decoder layer}
\label{app:stiefel}

Let the first decoder convolution have weight tensor
\[
W \in \R^{c_{\mathrm{out}}\times c_{\mathrm{in}}\times k_1\times k_2}.
\]
Flattening the input and spatial dimensions yields a matrix
\[
A \in \R^{c_{\mathrm{out}}\times p},
\qquad
p \coloneqq c_{\mathrm{in}}k_1k_2,
\qquad
A=\mathrm{reshape}(W).
\]
When $c_{\mathrm{out}}\ge p$, we project $A$ after each optimizer step onto the set of matrices with orthonormal columns:
\[
A_{\mathrm{proj}}
=
A\,(A^\top A)^{-1/2},
\]
so that
\[
A_{\mathrm{proj}}^\top A_{\mathrm{proj}} = \I_p.
\]
The inverse square root is computed by eigendecomposition. If
\[
A^\top A = Q\Lambda Q^\top,
\qquad
\Lambda=\mathrm{diag}(\lambda_i),
\]
then
\[
(A^\top A)^{-1/2}
=
Q\,\mathrm{diag}(\lambda_i^{-1/2})\,Q^\top,
\]
with eigenvalues clipped below by a small positive value for numerical stability.

This projection enforces orthonormal columns only in the selected layer. It therefore improves conditioning of that layer, but does not imply near-isometry or curvature control for the full nonlinear decoder.

\paragraph{Fallback case.}
If $c_{\mathrm{out}}<p$ (not used in our main setup), the implementation falls back to row-wise normalization rather than a true Stiefel projection.

\section{Full-order dataset generation}
\label{app:dataset_generation}

We consider a parametric advection--diffusion--reaction (ADR) problem on the unit square
\[
\Omega = [0,1]^2,
\]
with state variable $u=u(x,t)$ and parameter vector
\[
\mu = (\mu_1,\mu_2,\mu_3)\in\R^3.
\]
In the implementation, $\mu_1$ is the diffusion coefficient, while $(\mu_2,\mu_3)$ specify the center of a Gaussian source term. 
The reaction coefficient is fixed to
\[
c=1,
\]
and the time-dependent advection field is
\[
b(t) = (\cos t,\sin t).
\]
The source term is chosen as
\[
f(x;\mu_2,\mu_3)
=
10\exp\!\left(
-\frac{(x_1-\mu_2)^2 + (x_2-\mu_3)^2}{0.072}
\right).
\]
The initial condition is
\[
u(x,0)=0.
\]

\paragraph{Spatial discretization.}
We solve the full-order model using \texttt{dolfinx} with first-order Lagrange finite elements on a uniform triangular mesh with $31\times 31$ subdivisions. 
This yields $32\times 32=1024$ scalar degrees of freedom, so each snapshot can be reshaped as a tensor in
\[
\R^{1\times 32\times 32}.
\]

\paragraph{Temporal discretization.}
The full-order trajectories are computed on the time interval
\[
[0,T],
\qquad
T=10\pi,
\]
using a backward Euler time discretization with
\[
N_t^{\mathrm{FOM}} = 1000
\]
time steps. 
Hence the fine-grid time step is
\[
\Delta t_{\mathrm{FOM}} = \frac{T}{1000},
\]
and each trajectory contains
\[
N_t^{\mathrm{FOM}}+1 = 1001
\]
stored snapshots, including both $t=0$ and $t=T$.

For reduced-model training, we construct a coarse temporal grid by subsampling every fifth fine-grid step. 
Therefore,
\[
\Delta t_{\mathrm{train}} = 5\,\Delta t_{\mathrm{FOM}},
\]
and each trajectory contains
\[
N_t^{\mathrm{train}}+1 = 201
\]
coarse snapshots.

\paragraph{Time splits.}
To distinguish training and long-horizon evaluation regimes, we define
\[
T_1 = 4\pi,
\qquad
T_2 = 5\pi.
\]
These induce the following temporal partition:
\begin{itemize}
    \item $[0,T_1]$: primary training portion;
    \item $(T_1,T_2]$: interpolation-in-time portion;
    \item $(T_2,T]$: extrapolation-in-time portion.
\end{itemize}
During reduced-model training, trajectories are sampled on the coarse grid, while long-horizon rollout evaluation is also reported on the fine grid.

\paragraph{Parameter domain and dataset splits.}
The full parameter domain is
\[
\mathcal P
=
[10^{-2},\,6\cdot 10^{-2}]
\times
[0.3,\,0.7]^2.
\]
Training parameters are sampled from the inner box
\[
\mathcal P_{\mathrm{train}}
=
[2\cdot 10^{-2},\,5\cdot 10^{-2}]
\times
[0.4,\,0.6]^2.
\]
More precisely, we form a tensor grid with $10$ points per parameter dimension, yielding
\[
10^3 = 1000
\]
parameter triples. 
These are randomly split into $80\%$ training and $20\%$ validation parameter sets.

To assess interpolation in parameter space, we use the tensor grid of coordinate-wise midpoints of the training grid, which gives
\[
9^3 = 729
\]
interpolation test parameters inside the same inner box.

To assess extrapolation, we sample
\[
200
\]
parameters uniformly from $\mathcal P\setminus \mathcal P_{\mathrm{train}}$, i.e., from the outer box while excluding the training box.

\paragraph{Stored trajectories.}
For each sampled parameter, we store both the fine-grid trajectory
\[
u_{\mathrm{fine}} \in \R^{1001\times 32\times 32}
\]
and the coarse-grid trajectory
\[
u_{\mathrm{coarse}} \in \R^{201\times 32\times 32}.
\]
The resulting dataset is used for snapshot-wise autoencoder pre-training and for subsequent latent-dynamics training and rollout evaluation.

\section{Evaluation protocol}
\label{app:eval_protocol}

We evaluate all methods on a common set of sampled rollout windows, using identical windows across methods in order to enable paired comparisons.

\paragraph{Evaluation split and horizon.}
Let \texttt{split} denote one of the dataset partitions (\texttt{train}, \texttt{interp}, or \texttt{extrap}) and let \texttt{field} denote the temporal grid used for evaluation (\texttt{u\_fine} or \texttt{u\_coarse}). 
In our main experiments, we evaluate on the fine grid (\texttt{u\_fine}) with rollout horizon
\[
H \in \{80, 160, 240, 320\},
\]
where the horizon is measured in indices of the chosen evaluation grid.

For the selected split and field, we first determine the admissible time segment from the HDF5 file. 
Rollout windows are then sampled only from this segment, so that every start index $k_0$ satisfies
\[
k_0 + H \le k_{\max},
\]
where $k_{\max}$ is the last admissible time index of the segment.

\paragraph{Sampling of rollout windows.}
We sample a finite collection of evaluation windows
\[
(\mu_j, k_{0,j}), \qquad j=1,\dots,N_{\mathrm{win}},
\]
where $\mu_j$ indexes a parameter instance and $k_{0,j}$ is the starting time index of the rollout window. 
In the default configuration, we sample
\[
N_\mu = 100
\]
parameter indices (with replacement), and for each sampled parameter we draw
\[
N_{\mathrm{starts}} = 10
\]
start times uniformly from the admissible segment. 
This yields
\[
N_{\mathrm{win}} = N_\mu \cdot N_{\mathrm{starts}} = 1000
\]
sampled windows in total.

To preserve exact pairing even when repeated $(\mu,k_0)$ pairs occur, each sampled window is assigned a unique identifier
\[
\texttt{window\_id}\in\{0,\dots,N_{\mathrm{win}}-1\}.
\]
This identifier, rather than the tuple $(\mu,k_0)$ alone, is used for paired comparisons and multi-run aggregation.

\paragraph{Normalization.}
All rollouts are evaluated using the same normalization constants $(u_{\min},u_{\max})$, computed once from the training split on the coarse temporal grid over training parameters and training-time indices only. 
These normalization constants are shared across all methods.

\paragraph{Window rollout.}
For a sampled window $(\mu,k_0)$, let
\[
u_0,u_1,\dots,u_H
\]
denote the corresponding full-order states on the chosen evaluation grid, and let
\[
t_0,t_1,\dots,t_H
\]
be their absolute times. 
Absolute times are used explicitly because the latent ODE depends on time. 
The evaluation pipeline is:
\begin{enumerate}
    \item normalize the initial state $u_0$ and encode it to obtain the latent initial condition
    \[
    z_0 = E_\phi(u_0);
    \]
    \item integrate the latent ODE
    \[
    \dot z(t) = f_\theta(t,z(t),\mu)
    \]
    on the absolute-time grid $(t_0,\dots,t_H)$ using the chosen explicit Runge--Kutta integrator;
    \item decode the latent predictions to obtain
    \[
    \hat u_i = D_\psi(z(t_i)), \qquad i=0,\dots,H;
    \]
    \item compare the decoded trajectory with the full-order trajectory in physical space.
\end{enumerate}

\paragraph{Rollout error metrics.}
Let
\[
e_i = \hat u_i - u_i,
\qquad
\|e_i\| = \|\mathrm{vec}(e_i)\|_2,
\qquad
\|u_i\| = \|\mathrm{vec}(u_i)\|_2.
\]
All summary metrics are computed after discarding the trivial initial point $i=0$, since $\hat u_0$ is obtained directly from the encoded initial condition. 
Thus, errors are aggregated over
\[
i=1,\dots,H.
\]

For each window, we report:
\[
\mathrm{abs\_mean} = \frac{1}{H}\sum_{i=1}^H \|e_i\|,
\qquad
\mathrm{abs\_max} = \max_{1\le i\le H}\|e_i\|,
\qquad
\mathrm{abs\_fin} = \|e_H\|,
\]
and
\[
\mathrm{rel\_mean} = \frac{1}{H}\sum_{i=1}^H \frac{\|e_i\|}{\|u_i\|+\epsilon},
\qquad
\mathrm{rel\_max} = \max_{1\le i\le H}\frac{\|e_i\|}{\|u_i\|+\epsilon},
\qquad
\mathrm{rel\_fin} = \frac{\|e_H\|}{\|u_H\|+\epsilon},
\]
with a small numerical constant $\epsilon=10^{-8}$ in the denominator. 
We also record the final-time mean-squared error
\[
\mathrm{mse\_fin}
=
\frac{1}{n}\,
\|\mathrm{vec}(e_H)\|_2^2.
\]

\paragraph{Intrinsic diagnostics.}
In addition to rollout errors, we compute local intrinsic diagnostics on a subset of evaluation windows. 
By default, we use the first
\[
N_{\mathrm{intr}} = 80
\]
sampled windows (identified by \texttt{window\_id}) and evaluate diagnostics at
\[
N_{\mathrm{steps}} = 6
\]
approximately evenly spaced time indices in the rollout window.

For each selected window and intrinsic time index, we record:
\begin{itemize}
    \item the singular values of the Jacobian of the latent dynamics vector field with respect to the latent state,
    \[
    J_f(t,z,\mu) = \frac{\partial f_\theta(t,z,\mu)}{\partial z},
    \]
    computed exactly by autodiff;
    \item the corresponding local dynamics diagnostics
    \[
    \mathrm{dyn\_sigma\_max} = \sigma_{\max}(J_f),
    \qquad
    \mathrm{dyn\_sigma\_min} = \sigma_{\min}(J_f),
    \]
    \[
    \mathrm{dyn\_cond} = \frac{\sigma_{\max}(J_f)}{\sigma_{\min}(J_f)+10^{-12}};
    \]
    \item a decoder-gain proxy
    \[
    \mathrm{dec\_gain} \approx \|J_D(z)\|_2,
    \]
    computed by power iteration on $J_D(z)^\top J_D(z)$;
    \item the latent tracking error
    \[
    \mathrm{latent\_err}
    =
    \|z_{\mathrm{pred}} - E_\phi(u_{\mathrm{true}})\|_2,
    \]
    i.e., the distance between the predicted latent state and the AE-encoded ground-truth state at the same time.
\end{itemize}

\paragraph{Deterministic decoder-gain estimate.}
To avoid additional Monte Carlo variance in the intrinsic diagnostics, the decoder-gain power iteration uses a fixed probe initialization (or an equivalent deterministic seeding rule). 
This makes \texttt{dec\_gain} reproducible across repeated evaluations and directly comparable across methods.

\paragraph{Per-method summaries.}
For the main reported statistics, rollout errors are first averaged within each matched autoencoder/NODE realization and are then aggregated across realizations. Window-level aggregation is used only within a run.

Intrinsic diagnostics are aggregated separately over the selected intrinsic windows and intrinsic time indices.

When multiple evaluation runs are concatenated, the saved tables include both the method label and the run identifier:
\[
(\texttt{method},\texttt{run},\texttt{window\_id}).
\]
This ensures that summaries remain correctly indexed by method and that multi-run aggregation does not merge distinct runs or repeated windows.

\paragraph{Paired comparisons.}
To compare a method against the unregularized baseline, we perform paired tests using the exact same sampled windows. 
Pairing is carried out using the unique window identifier (and the run identifier when multiple runs are pooled), so repeated $(\mu,k_0)$ pairs remain distinct observations. 
For a chosen scalar metric (in our main experiments, \texttt{rel\_mean}), we compute per-window differences relative to the baseline, average them within matched autoencoder/NODE realization per method and report:
\begin{itemize}
    \item the mean of the paired differences;
    \item a one-sided Wilcoxon signed-rank $p$-value testing whether the method improves over the baseline.
\end{itemize}

This protocol ensures that all method comparisons are performed on the same rollout windows, with no loss of duplicated samples and with correct bookkeeping across repeated NODE runs.

\section{Additional results}

Additional rollout experiments were performed with NODE checkpoint selection using a shared validation-rollout target, defined post-hoc. 

The goal of these experiments was to determine whether differences between methods during rollouts on \texttt{extrap} split can be mitigated by matching performance on \texttt{train}/\texttt{val} rollouts.

We show in Table~\ref{tab:paired_target} that the overall pattern is the same: NODE seeds trained with AE regularized via Stiefel projection perform slightly better vanilla baseline, whereas other regularization methods show higher rollout errors.

\begin{table}[h]
\caption{Paired rollout-error differences relative to the vanilla baseline}
\label{tab:paired_target}
\smallskip
\begin{tabular}{lrrrr}
\toprule
Method & $\Delta \varepsilon_{\text{max}}$ & Std & 95\% CI & $p$-value \\
\midrule
Stiefel projection & 0.009 & 0.019 & 0.010 & 0.058 \\
Isometry penalty   & -0.012 & 0.020 & 0.010 & 0.988 \\
Gain penalty       & -0.014 & 0.018 & 0.010 & 0.997 \\
Curvature penalty  & -0.016 & 0.020 & 0.011 & 0.997 \\
\midrule
\midrule
Method & $\Delta \varepsilon_{\text{mean}}$ & Std & 95\% CI & $p$-value \\
\midrule
Stiefel projection & 0.005 & 0.012 & 0.006 & 0.080 \\
Isometry penalty   & -0.015 & 0.012 & 0.006 & 1.000 \\
Gain penalty       & -0.016 & 0.011 & 0.006 & 1.000 \\
Curvature penalty  & -0.018 & 0.012 & 0.007 & 1.000 \\
\bottomrule
\end{tabular}

\smallskip
\textit{Paired differences are defined as $\Delta = \varepsilon(\texttt{vanilla})-\varepsilon(\texttt{method})$, so positive values indicate an improvement over the unregularized baseline. 
Values are averaged over matched AE/NODE realizations. 
Results are shown for rollout horizon $H=320$ on the fine grid, with NODE checkpoints selected using the shared validation-MSE target reached by all methods. 
Uncertainty is reported as a 95\% confidence interval. 
The reported $p$-values correspond to the one-sided Wilcoxon signed-rank test for the alternative hypothesis $\Delta>0$, i.e.\ that the method improves over the baseline.}
\end{table}

\end{document}